# Pricing Football Players using Neural Networks

Sourya Dey

## Abstract:

We designed a multilayer perceptron neural network to predict the price of a football (soccer) player using data on more than 15,000 players from the football simulation video game FIFA 2017. The network was optimized by experimenting with different activation functions, number of neurons and layers, learning rate and its decay, Nesterov momentum based stochastic gradient descent, L2 regularization, and early stopping. Simultaneous exploration of various aspects of neural network training is performed and their trade-offs are investigated. Our final model achieves a top-5 accuracy of 87.2% among 119 pricing categories, and places any footballer within 6.32% of his actual price on average.

## Introduction:

Football (or soccer, if you're in North America) is the most widespread team sport played in the world [1]. Apart from international football, the bulk of football matches take place at the level of domestic clubs. Most countries have an established football league where clubs representing different geographical regions compete against one another. There are no restrictions on the players who may represent a particular club. For example, the current starting lineup for Chelsea – a club based in London – has 2 players each from Brazil, Spain, France and Belgium, and 1 each from Nigeria, England and Serbia. One of the most intriguing aspects of football, and possibly the biggest headache of a football manager, is buying and selling players for appropriate prices. The price of a player is a function of his technical abilities such as ball control and tackling, but it also depends a lot on physical attributes such as speed and stamina due to the high energy levels required in the modern game. Factors such as the age of a player and reputation on the international stage also influence the price. Till date, there are no standardized means to price a football players and some star players such as Paul Pogba have recently switched clubs for prices in excess of $100 million [2]. The question which therefore arises is: what is the right price for a football player?

This project uses data from FIFA 2017 to construct a pricing model for football players. FIFA 2017 is a football simulation video game developed by Electronic Arts (EA) Sports, which has official licenses for most of the major footballing teams and players in the world [3]. The complete roster in the game includes 17240 players, each of whom has attributes as mentioned above and a preset price. All these attributes and prices are updated on a weekly basis based on real-life performances, which allows FIFA 2017 to keep its data fresh [4]. The initial data for each player is gathered from a team of 9000 data reviewers comprising managers, scouts and season ticket holders [5]. They rate each player in 300 categories and provide EA Sports with the data, which is then fed into a weighting algorithm to compute the final attributes for each player to be used in the game. In rare cases, EA Sports bumps up or down the rating of a certain player on a subjective basis.

This project builds a neural network which accepts player attributes as input and computes a price. A neural network is a machine learning technique where layers of neurons perform computations and update internal parameters based on training data. Supervised learning [6] with stochastic gradient descent [7] has been used for this project. There are 41 input features – 37 on a scale of 0-99, age on a scale of 16-43, and 3 on a scale of 1-5 stars [8]. Goalkeeping features have not been used and consequently, the model does not work for goalkeepers. There are 119 pricing categories, the lowest being $43,000 and the highest $36,700,000. The prices occurring in-game are quantized to specific values, for example, a few hundred players are priced at $1,100,000. Taking a cue from how the MNIST dataset [9] is split, 10,914 players have been used for training, 1926 for validation, and 2500 for test. This project experiments with different activation functions [10] for different network layers, the number of hidden layers and the number of neurons in each, appropriate values for learning rate and its annealing, weight regularization using L2 norm, stochastic gradient descent using different batch sizes, Nesterov momentum based parameter updates in gradient descent, and early stopping of training to prevent overfitting [11], [12]. The final top-5 accuracy obtained is 87.2%.

## Network Experiments:

*Activation Functions*: Any activation function can be used for the hidden layers, such as Rectified Linear Unit (ReLU), hyperbolic tangent (tanh) or sigmoid. The ideal output is one-hot encoded in 119 categories. So the activation function for the output layer needs to be between 0 and 1, leading to a choice between the squashing sigmoid function and the softmax probability density. Experiments led me to pick ReLU activation for all the hidden layers except for the output, which is softmax.

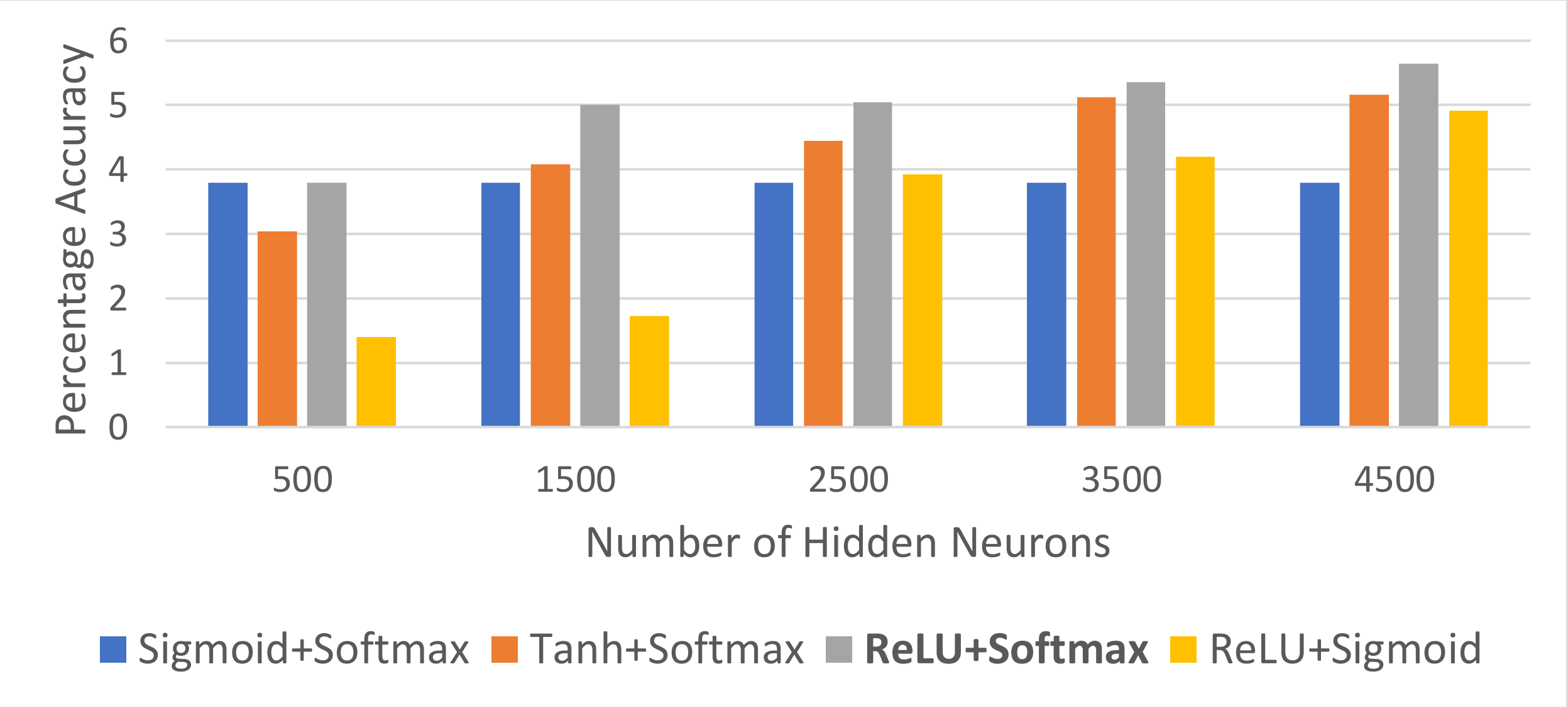

*Number of Neurons in 1st Hidden Layer*: A trend wasn't clearly discernible. Maximum accuracy is obtained for 3900 neurons, but after other parameters are varied, 2000 neurons is a better choice.

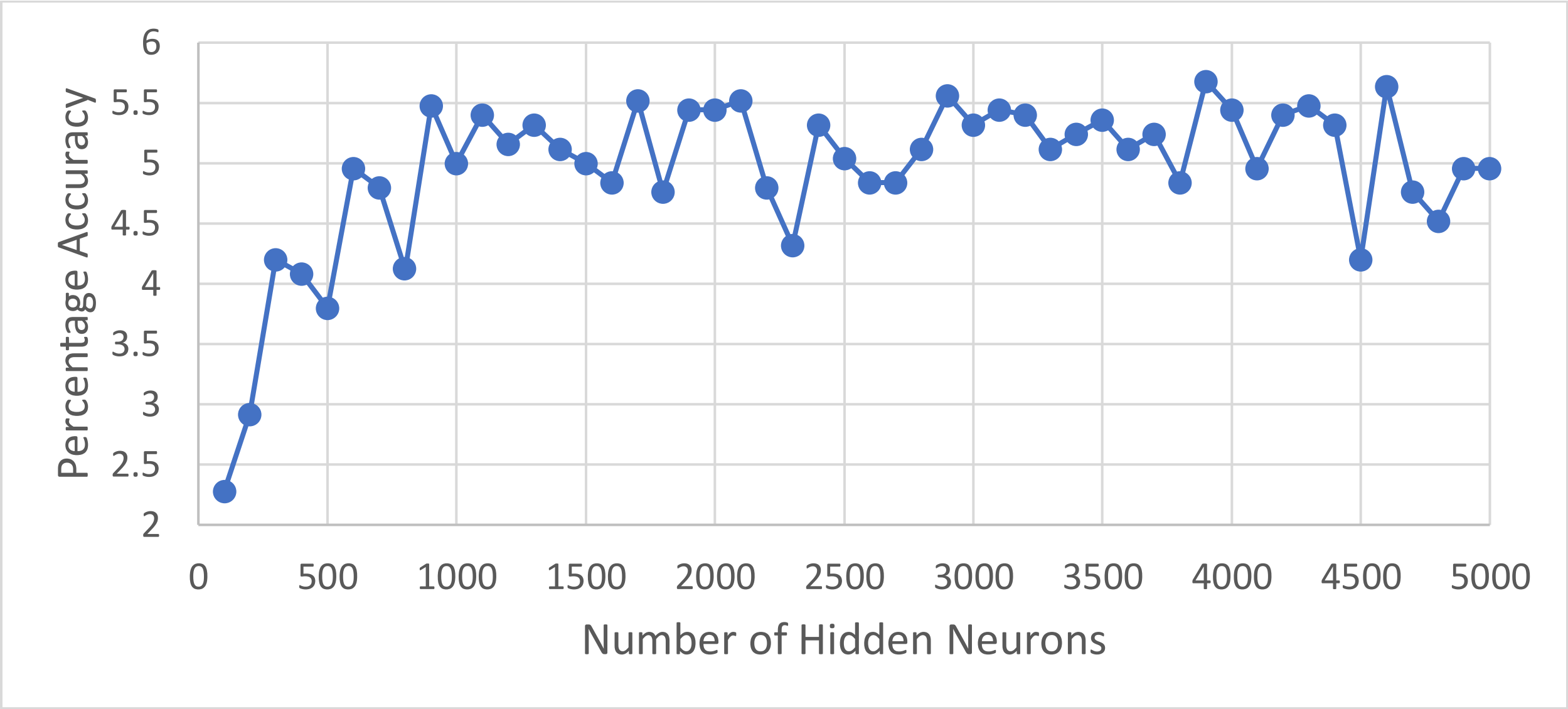

*Learning Rate*: I initially varied the learning rate logarithmically, then switched to linear variations to fine tune it. A value of 0.01 is chosen.

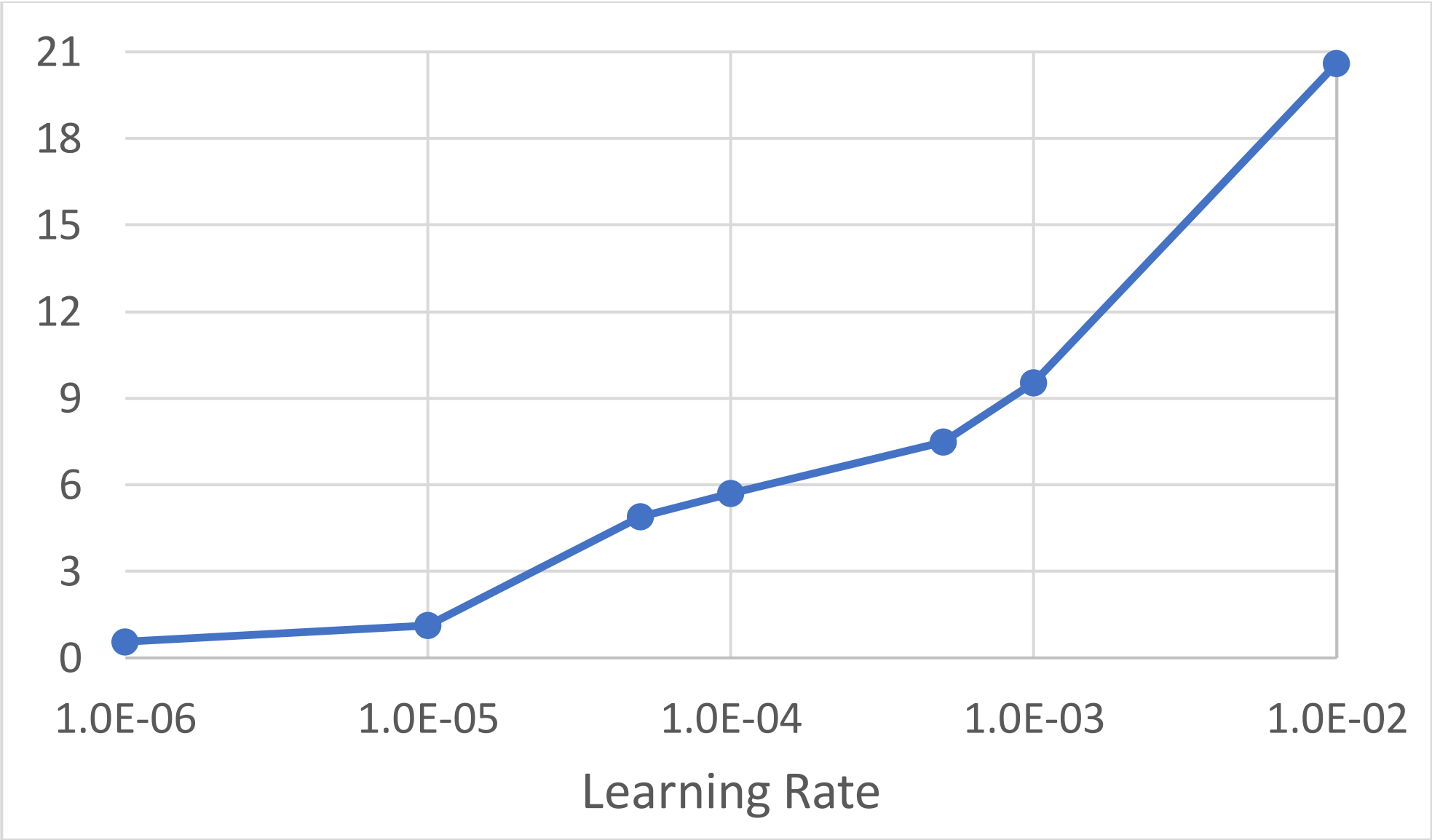

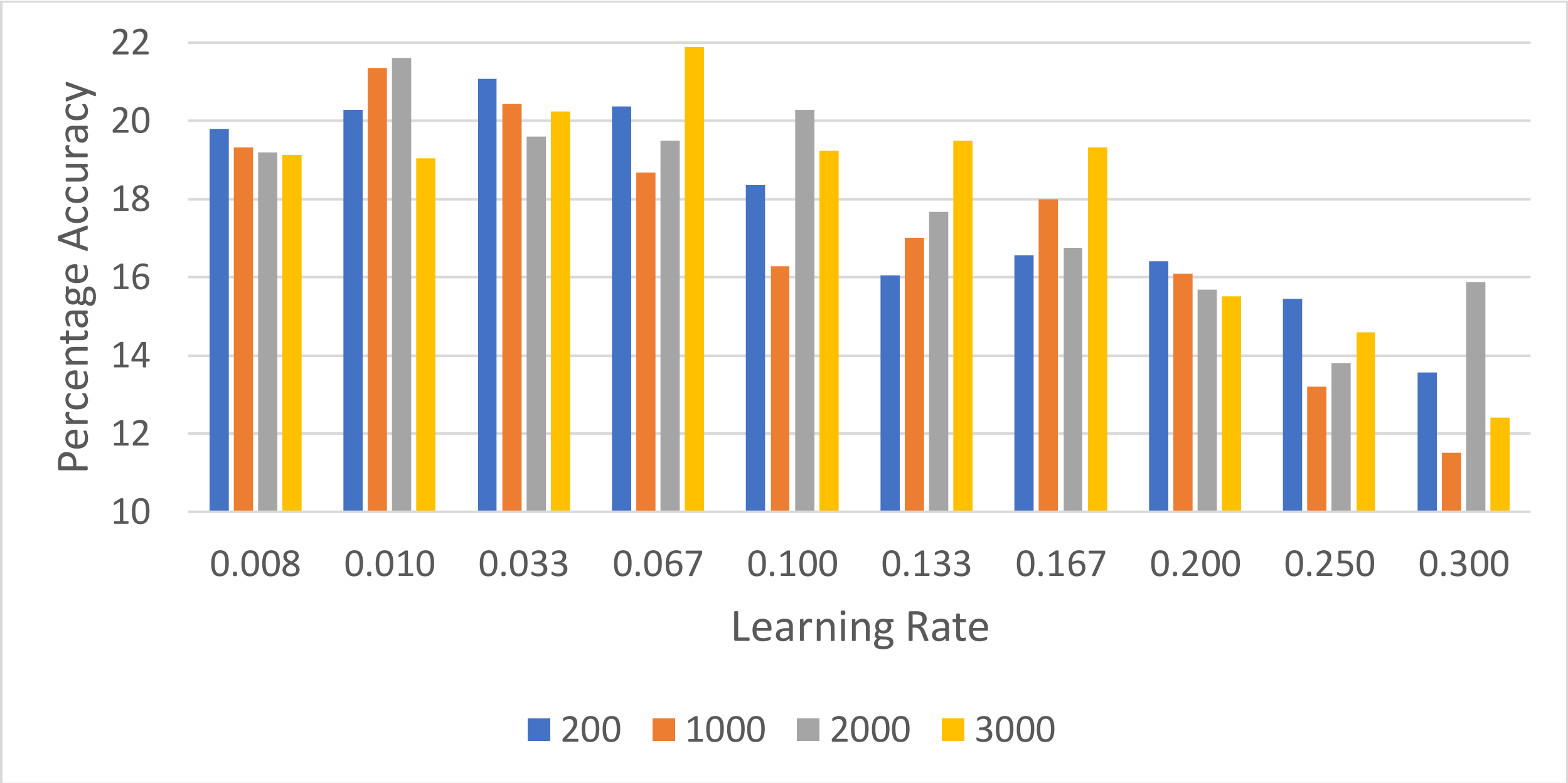

The color coded numbers (200,1000,2000,3000) are the number of neurons in the 2nd hidden layer.

*Number of Neurons in subsequent Hidden Layers*: No trend was clearly observed for the 2nd hidden layer. For the 3rd, less neurons gave better performance. The final numbers chose were 1500 and 500. This gives a final network configuration of [41,2000,1500,500,119].

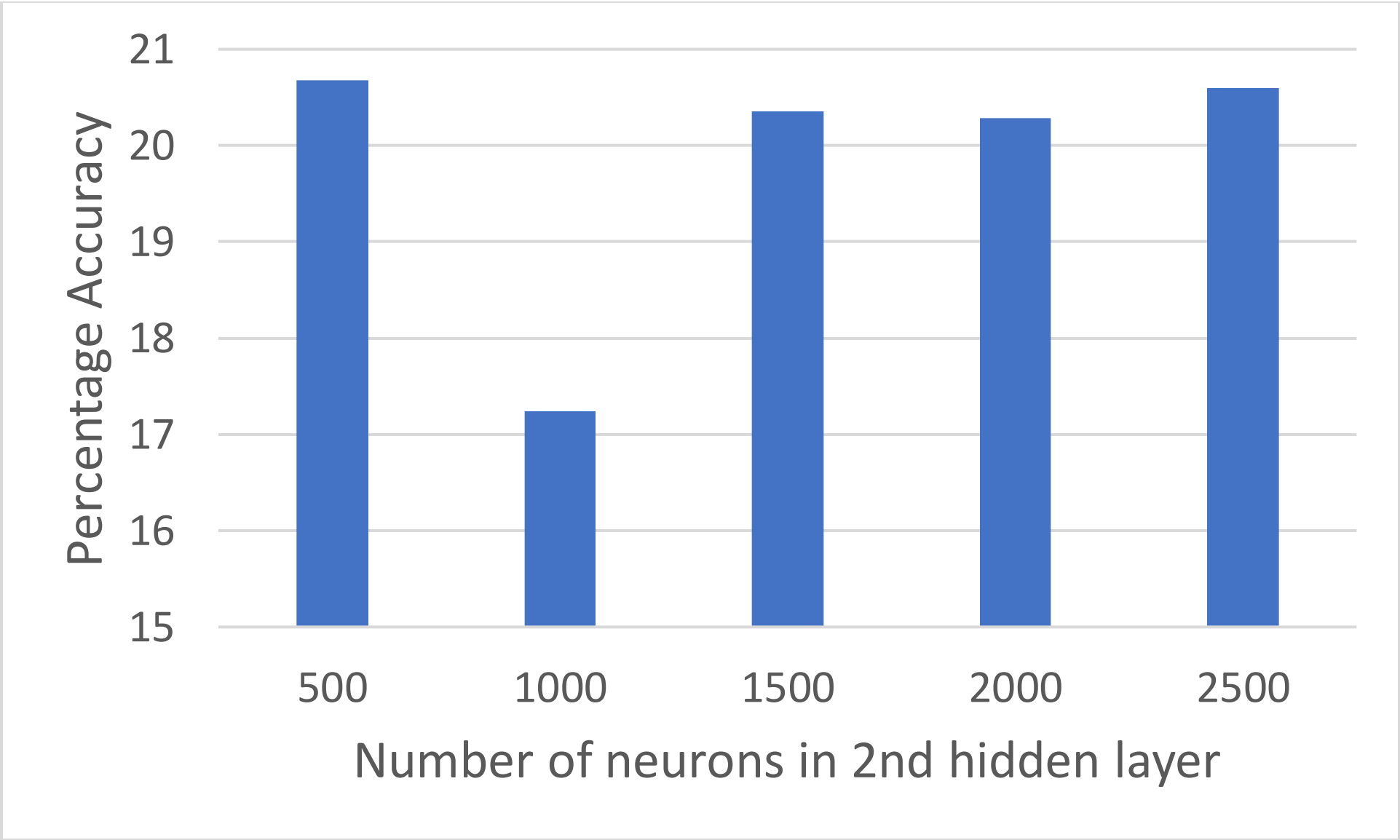

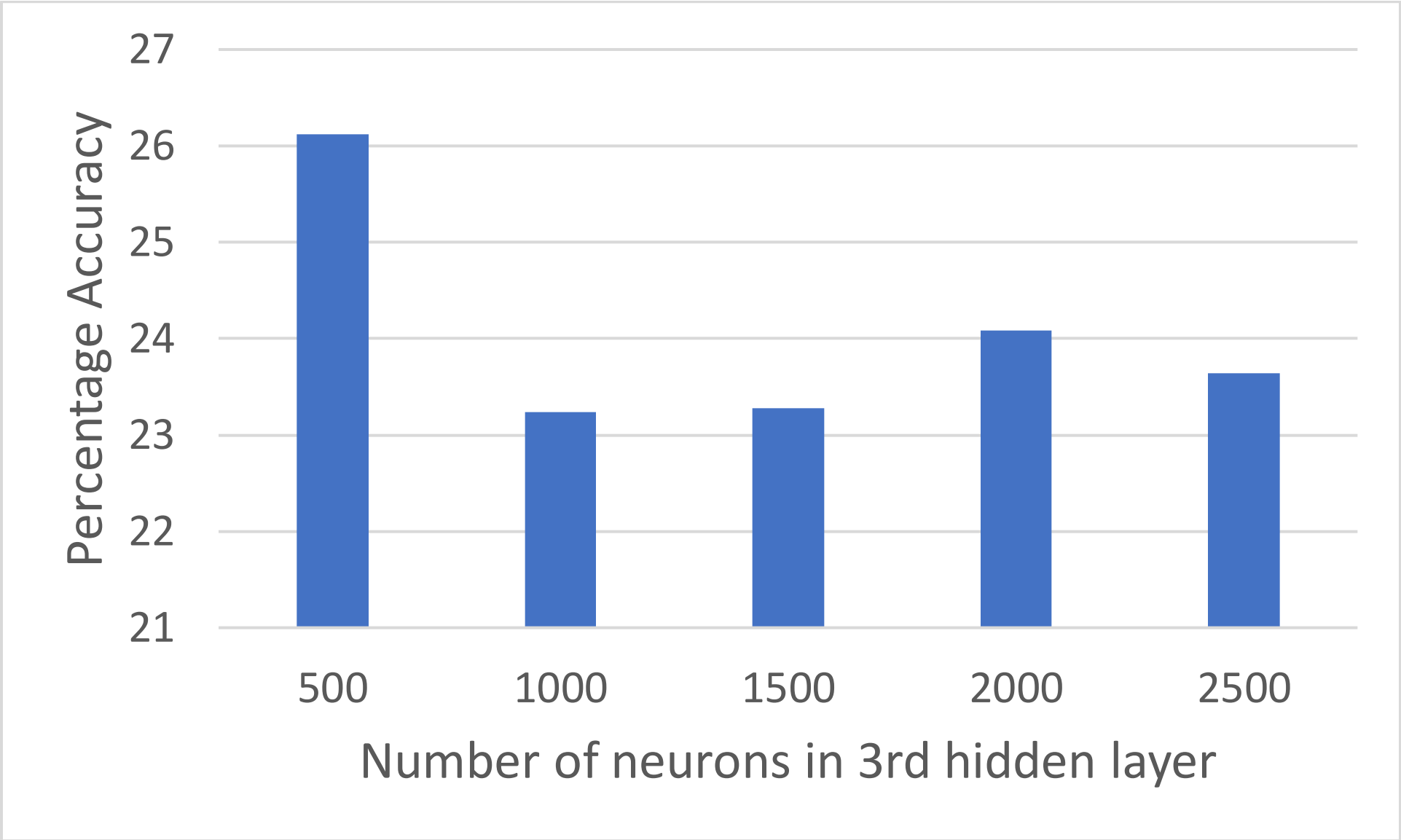

*Learning Rate Decay (Annealing)*: As the network learns, the cost function gets minimized according to the gradient descent algorithm. With a slow learning rate, the network takes a long time to learn. With a large learning rate, there is a danger of oscillating about the minimum point instead of settling in it. To mitigate the above issues, it is beneficial to pick a large learning rate to being with and reduce (anneal) it with every epoch, as shown [13]:

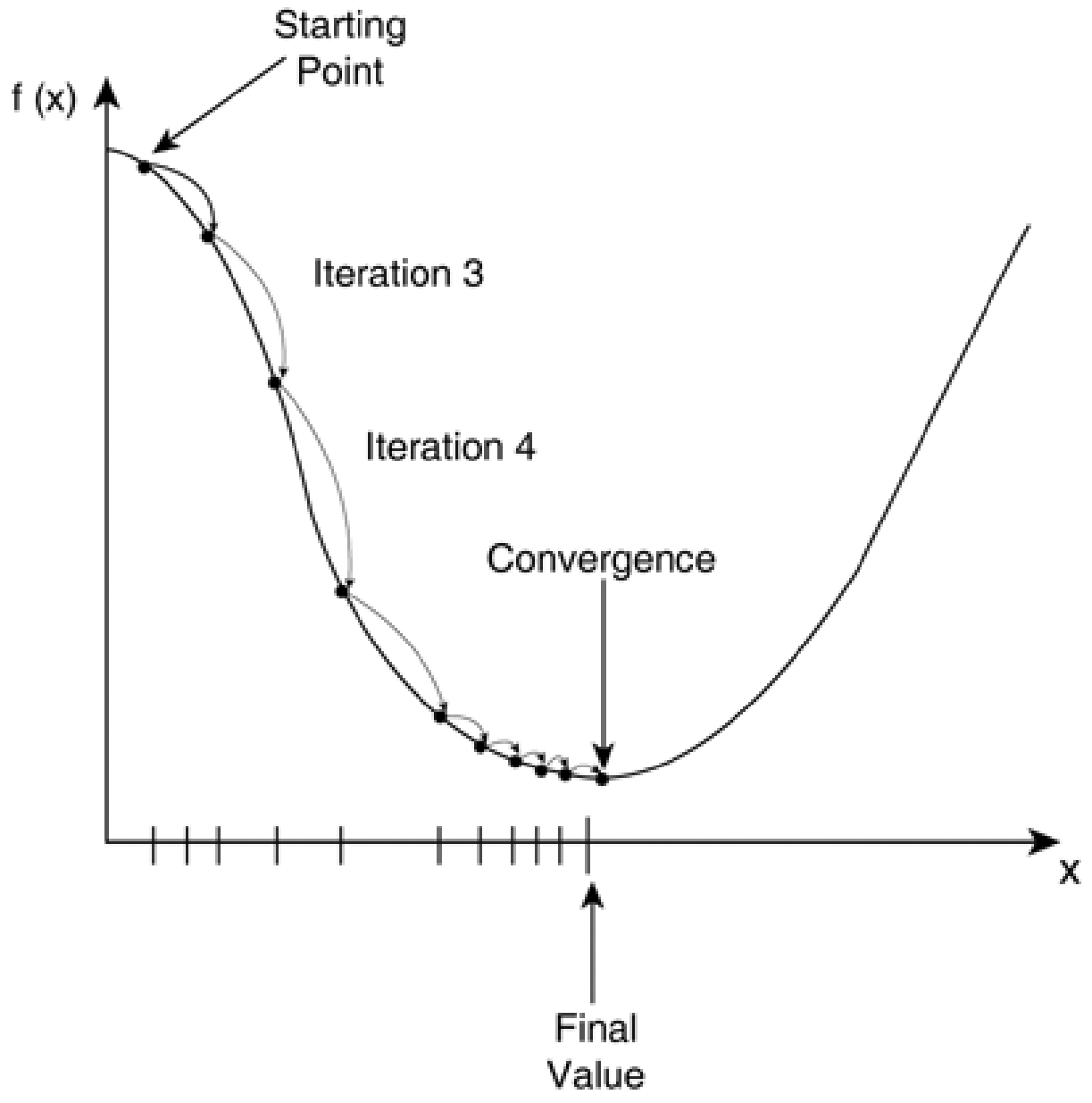

The rule followed is:

$$\eta_t = \frac{\eta_0}{1 + \boldsymbol{k}t}$$

where $\eta_0$ is the initial learning rate, $\eta_t$ is the learning rate after $t$ epochs, and $\boldsymbol{k}$ is the annealing coefficient. From experiments, I picked annealing coefficient = 0.001.

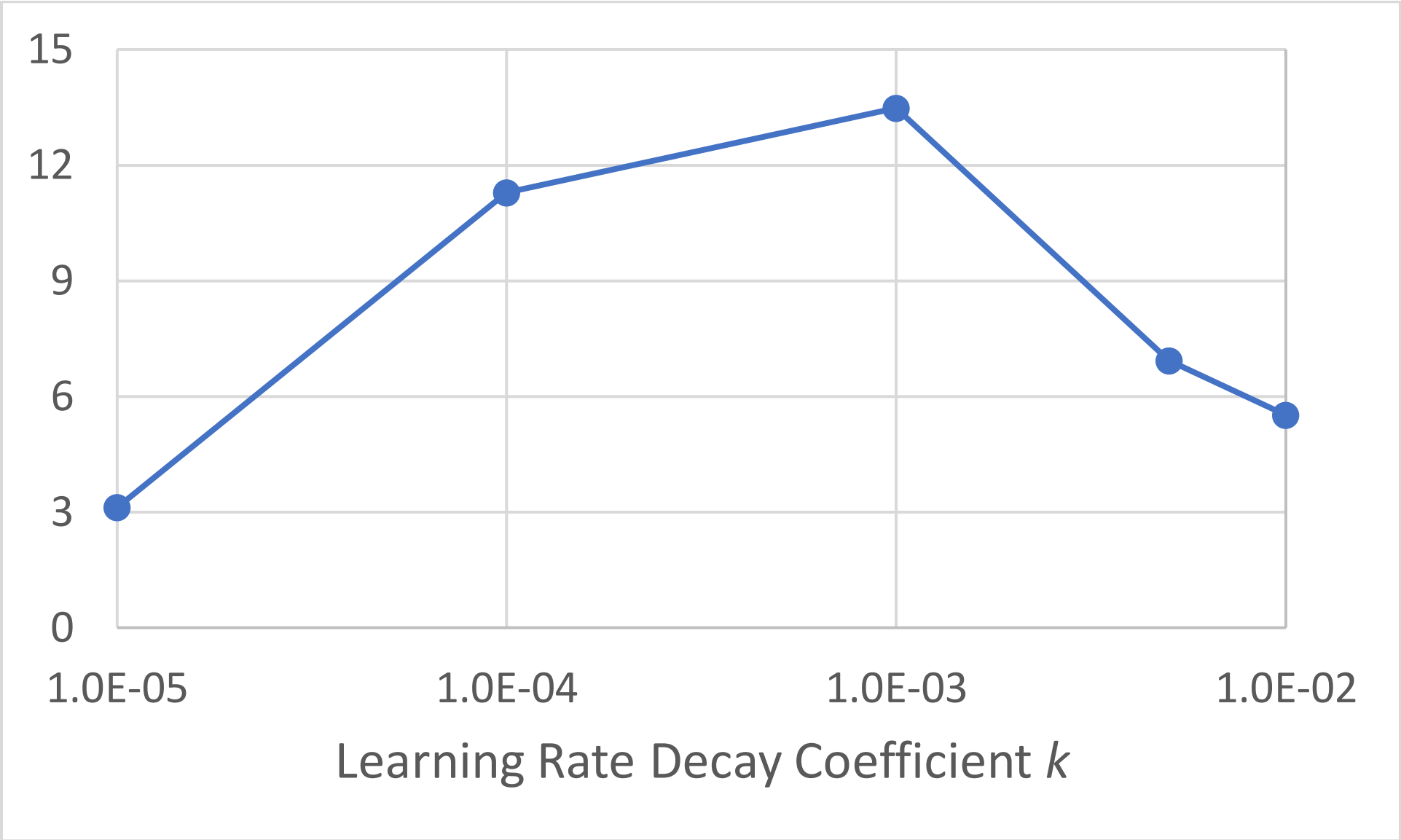

*Nesterov Momentum*: Ordinary gradient descent is akin to simple harmonic motion, where the restoring force on the pendulum is proportional to its position. The update is given as:

$$w_{t+1} = w_t - \eta \frac{\partial(Cost)}{\partial w}\bigg|_t$$

In reality, the motion is damped by air resistance proportional to the velocity of the pendulum. This analogy also applies to friction affecting a ball rolling down a hill. Then the update becomes:

$$w_{t+1} = (w + \boldsymbol{\mu}\Delta w)_t - \eta \frac{\partial(Cost)}{\partial(w + \boldsymbol{\mu}\Delta w)}\bigg|_t$$

This is shown in the left figure. On the other hand, Nesterov momentum updates first compute the new position of the ball and take the derivative with respect to that, as shown in the right figure [12].

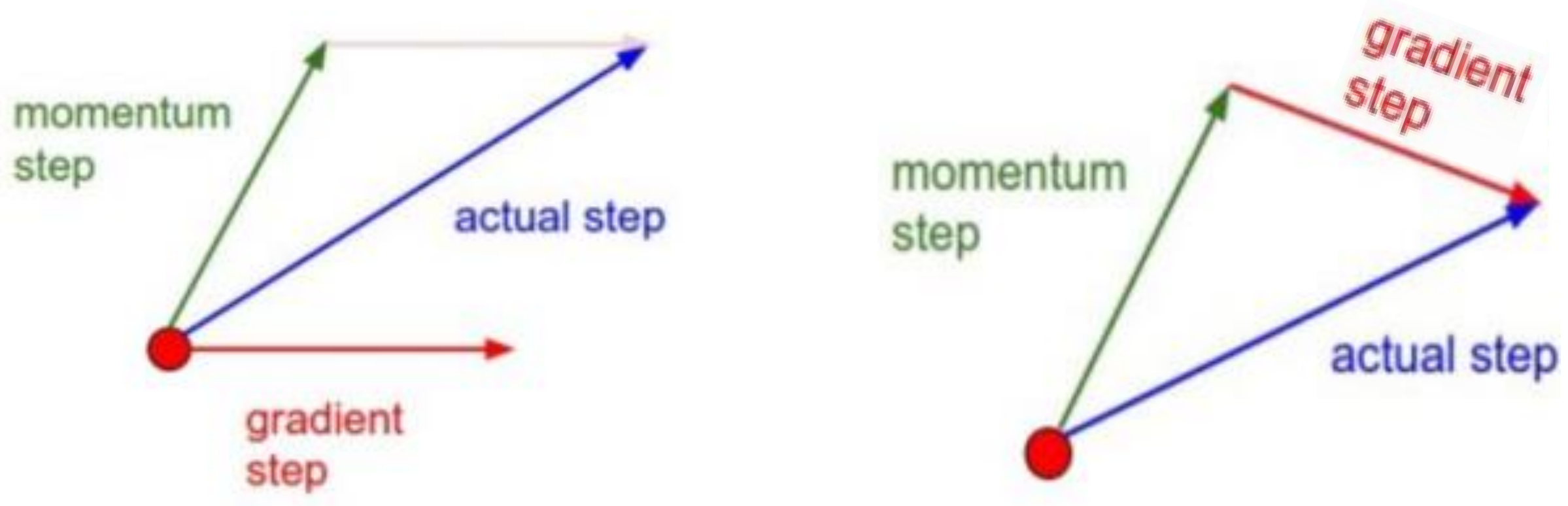

The corresponding update is:

$$w_{t+1} = (w + \boldsymbol{\mu}\Delta w)_t - \eta \frac{\partial(Cost)}{\partial(w + \boldsymbol{\mu}\Delta w)}\bigg|_t$$

where $\boldsymbol{\mu}$ is a hyperparameter which has to be less than 1. I picked a value of 0.99.

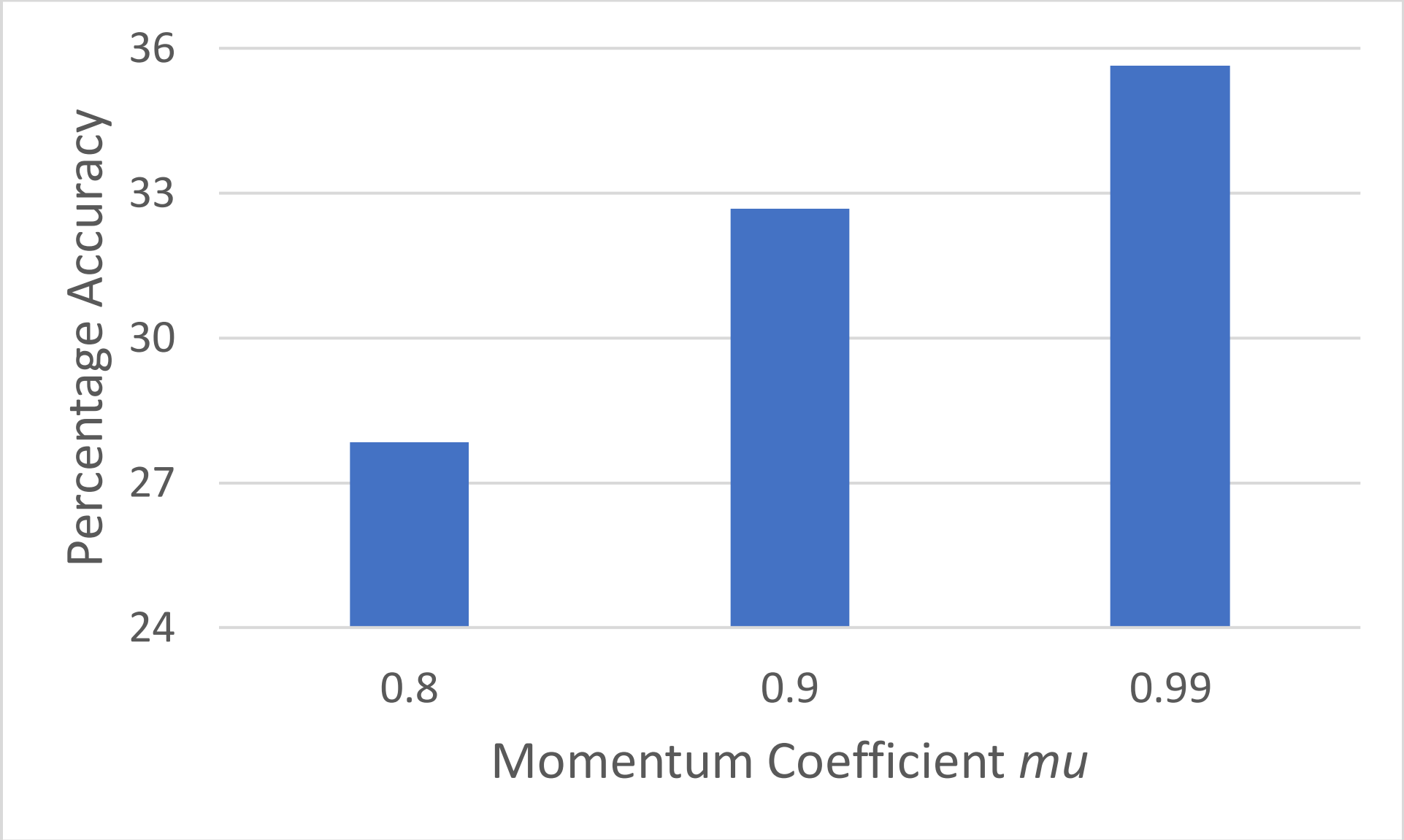

*L2 Regularization*: Performance of the network can be improved by penalizing high values of weights, so that no particular weight gets out of hand and adversely affects the network. This is done by adding the following term to the existing cost:

$$Extra\ Cost = \boldsymbol{\lambda} \sum w^2$$

I picked $\boldsymbol{\lambda} = 5 \times 10^{-4}$

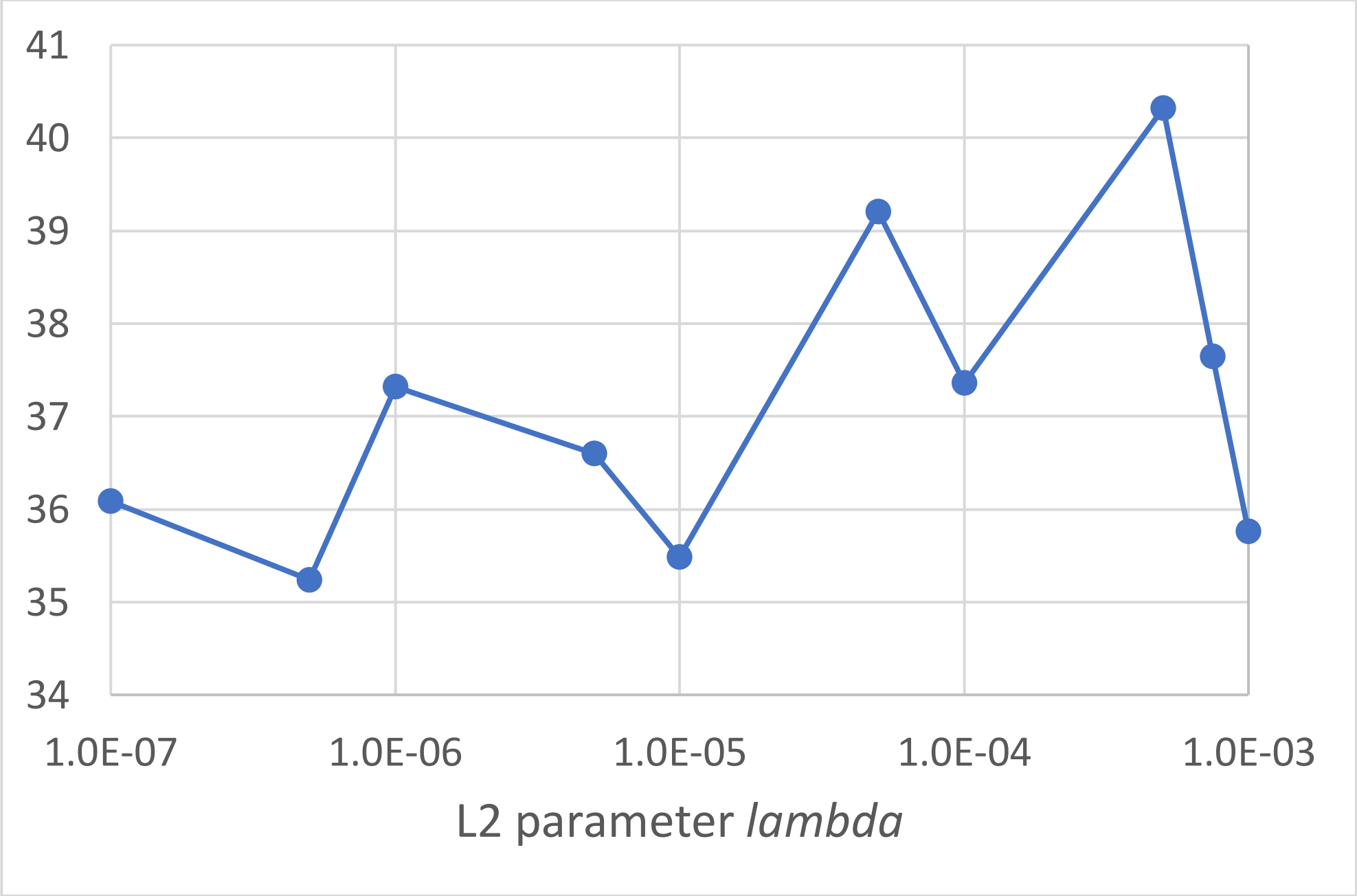

*Early Stopping*: Overtraining is a common issue in neural networks. There comes a point when the network is learning the specific data, it isn't learning general features any more. As a result, training accuracy keeps on improving, but validation and test accuracy suffers as shown [14]. Training should be stopped when validation accuracy doesn't increase for a certain number of epochs, which I picked to be 10.

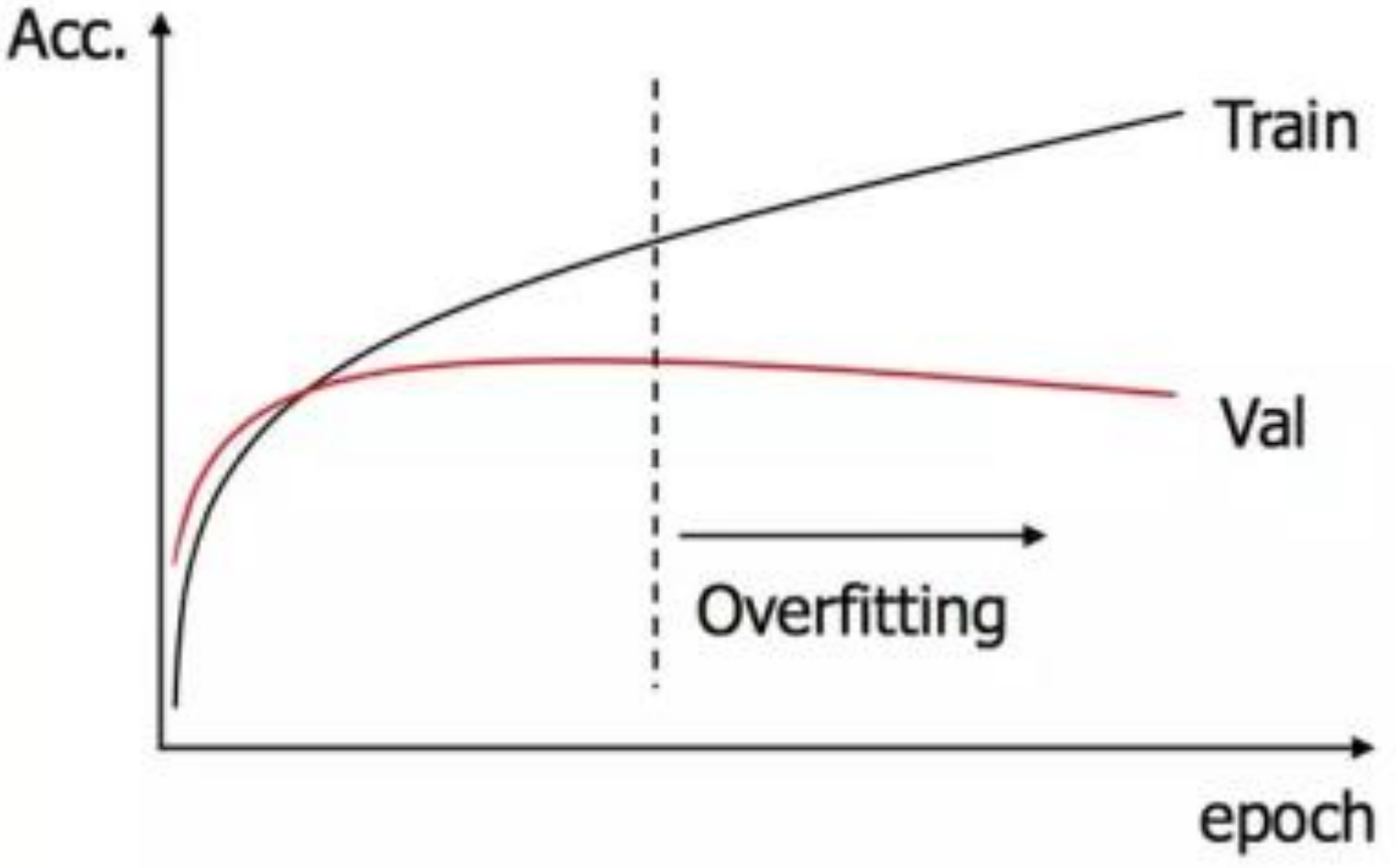

## Results:

The final network parameters chosen were:

- Network configuration = [41,2000,1500,500,119]
- ReLU activation for all hidden layers, finally softmax output
- L2 Regularization Coefficient = 0.0005
- Learning rate = 0.01, annealing coefficient = 0.001
- Nesterov momentum coefficient = 0.99
- Minibatch size for stochastic descent = 20

*Top-1 accuracy* (or simply, accuracy) was the metric chosen for network optimizations. This means that a test sample is correctly classified if the predicted output class matches exactly with the actual output class. Since there are 119 output classes, this metric fails to give excellent results. For an application such as pricing an item, it is more important to predict a price close to the actual value instead of getting an exact match. Note that the output neurons are arranged in ascending order of prices. This means that if the neuron for the predicted class is in close vicinity of the neuron for the actual class, the prediction is satisfactory. Based on this, *top-3 accuracy* and *top-5 accuracy* metrics can be defined where the predicted neuron is at an absolute distance no less than 1 and 2, respectively, from the actual neuron.

Another metric used was *Average Percentage Error* (APE) in price, defined as:

$$APE = \left( Avg_{All\ Test\ Samples} \frac{|True\ price - Predicted\ price|}{True\ price} \right) \times 100$$

The test results of the final trained network using these metrics are:

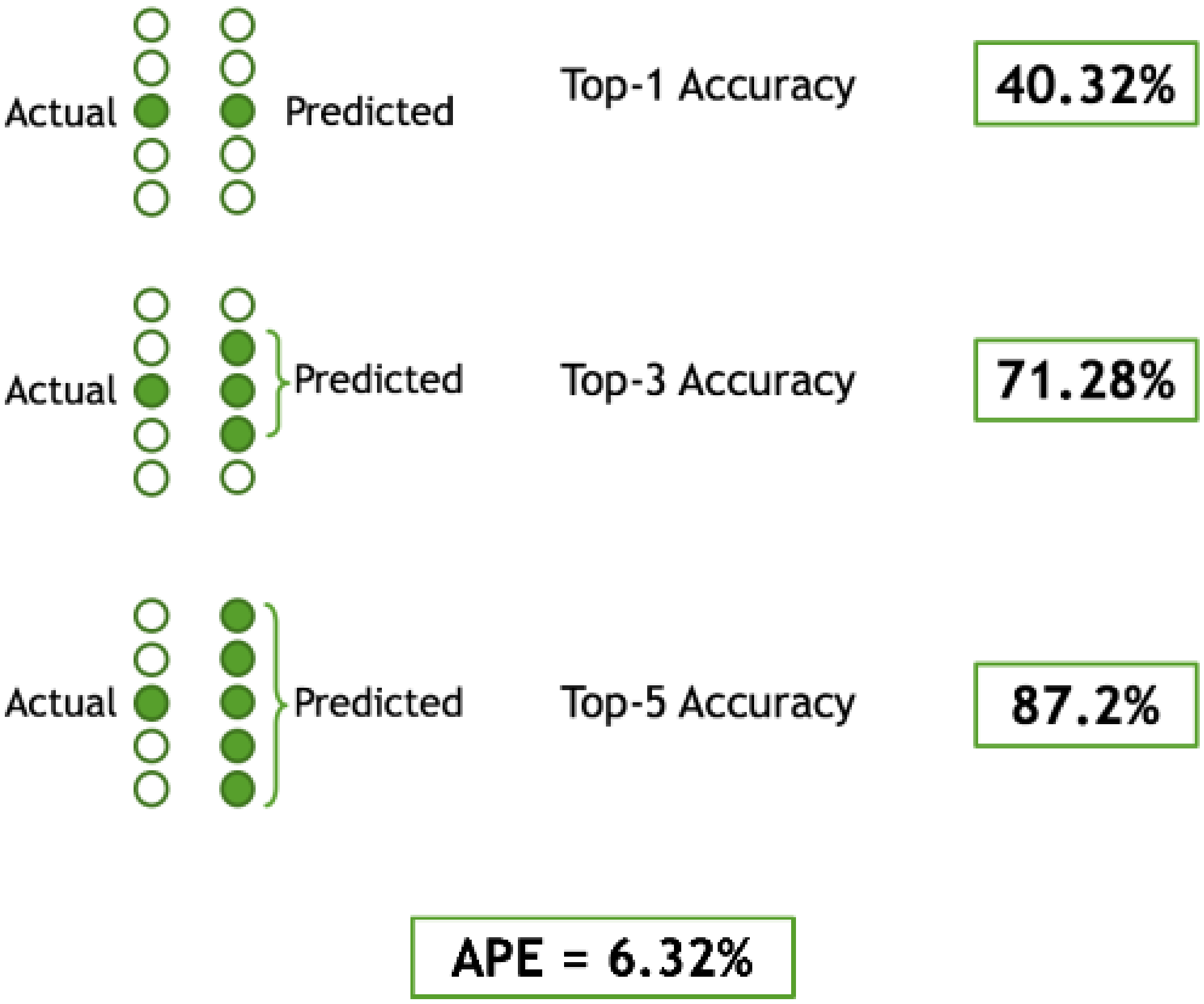

## Conclusion:

Neural networks are extremely versatile machine learning tools that can learn features and use this knowledge to make predictions. This project demonstrates their capability by solving a pertinent real-world problem – pricing football players in the transfer market – which is one of the leading contentious issues among football players, managers, agents, owners and, of course, fans, as they wait with bated breath for new talent to potentially arrive at their club before the transfer window closes. The neural network model used here utilizes several machine learning features such as regularization, annealing and momentum descent, and places a footballer within 6.32% of his actual price. The model does not take goalkeepers into account and fails to predict prices of outlying star players such as Lionel Messi, who has a price tag estimated to be well in excess of $100 million [15]. The problem of vanishing gradients in a deep network has not been investigated, which opens up the doors to potential improvements such as a using different learning rates for each layer.

## Technical Details:

The project uses Keras [16] – a machine learning library for Python, with Theano [17] as the backend. Complete datasets and code for this project are available at https://github.com/souryadey/footballer-price.git

## Acknowledgements:

I would like to acknowledge my professor Bart Kosko for giving me the opportunity to perform this project. I would also like to acknowledge the website sofifa.com for providing player attributes from FIFA 2017 in a user-friendly fashion. Finally, thanks to Nitin Kamra for allowing me to use some of his code written for CSCI567 'Machine Learning' – a course offered by USC in Fall 2016.